%% file: 00-article.tex
\documentclass[letterpaper, 10 pt,twoside]{IEEEtran}  

\IEEEoverridecommandlockouts                              

\usepackage[utf8]{inputenc}
\usepackage{booktabs}
\usepackage{multirow}
\usepackage{nicematrix}
\usepackage{xcolor}
\usepackage{changepage}
\usepackage{float}
\usepackage{subcaption}
\usepackage{algorithm} 
\usepackage{algpseudocode}
\usepackage{adjustbox}

\usepackage{graphicx}
\usepackage{amsbsy}
\usepackage{amsmath}
\usepackage{xcolor}
\usepackage{array}
\usepackage{longtable}
\usepackage{calc}
\usepackage{subfiles}
\usepackage{comment}

\makeatletter
\let\NAT@parse\undefined
\makeatother
\usepackage[pagebackref=true,colorlinks=true,citecolor=blue,linkcolor=blue]{hyperref} 

\title{\LARGE \bf
Real3D-Aug: Point Cloud Augmentation by Placing Real Objects with Occlusion Handling for 3D Detection and Segmentation

}

\author{Petr Šebek$^{*}$, Šimon Pokorný$^{*}$, Patrik Vacek, Tomáš Svoboda \\
Dept. of Cybernetics, Faculty of Electrical Engineering, Czech Technical University in Prague
\thanks{$*$Both authors contributed equally.}%
\thanks{Corresponding author: {\tt\small vacekpa2@fel.cvut.cz}}%
}

\makeatletter
\def\ps@IEEEtitlepagestyle{%
  \def\@oddfoot{\mycopyrightnotice}%
  \def\@oddhead{\hbox{}\@IEEEheaderstyle\leftmark\hfil\thepage}\relax
  \def\@evenhead{\@IEEEheaderstyle\thepage\hfil\leftmark\hbox{}}\relax
  \def\@evenfoot{}%
}
\def\mycopyrightnotice{%
  \begin{minipage}{\textwidth}
  \centering \scriptsize
  This work has been submitted to the IEEE for possible publication. Copyright may be transferred without notice, after which this version may no longer be accessible.
  \end{minipage}
}
\makeatother

\begin{document}

\maketitle


\subfile{01-abstract}

\subfile{02-introduction}

\subfile{03-related_work}


\subfile{04-method}

\subfile{05-experiments}

\subfile{07-conclusion}

\section*{ACKNOWLEDGMENT}
This work was supported in part by OP VVV MEYS funded project CZ.02.1.01/0.0/0.0/16\_019/0000765 ``Research Center for Informatics'', and by Grant Agency of the CTU Prague under Project SGS22/111/OHK3/2T/13. Authors want to thank colleagues from Valeo R\&D for discussions and Valeo company for a support.

\bibliographystyle{plain}
\bibliography{bibliography}

\addtolength{\textheight}{-12cm}   





\end{document}

%% file: 01-abstract.tex
\begin{abstract}

Object detection and semantic segmentation with the 3D lidar point cloud data require expensive annotation. We propose a data augmentation method that takes advantage of already annotated data multiple times.
We propose an augmentation framework that reuses real data, automatically finds suitable placements in the scene to be augmented, and handles occlusions explicitly. Due to the usage of the real data, the scan points of newly inserted objects in augmentation sustain the physical characteristics of the lidar, such as intensity and raydrop.
%
%
%
The pipeline proves competitive in training top-performing models for 3D object detection and semantic segmentation. 
The new augmentation provides a significant performance gain in rare and essential classes, notably 6.65$\%$ average precision gain for ``Hard'' pedestrian class in KITTI object detection or 2.14 mean IoU gain in the SemanticKITTI segmentation challenge over the state of the art.

Codes are available at \url{https://github.com/ctu-vras/pcl-augmentation}.


\end{abstract}

%% file: 02-introduction.tex
\section{INTRODUCTION}

Accurate detection and scene segmentation are integral to any autonomous robotic pipeline. Perception and understanding is possible thanks to various sensors, such as RGB cameras, radars, and lidars. 
These sensors produce structural data and must be interpreted for the proper function of critical safety systems.
We focus on lidars. Recently, the most promising way to process lidar data is to train deep neural networks \cite{cite:PointPillars,cite:3Dcylinder,cite:SPVNAS} with full supervision, which require a large amount of annotated data.

The manual annotation process is very time and resource consuming. For example, to perform semantic segmentation on lidar point clouds, one needs to accurately label all the points in the scene as a specific object class.
As a result, there is not enough annotated data to train large neural networks. 
Data augmentation is a way to effectively decrease the need for more annotated data by enriching the training set with computed variations of the data. This type of augmentation is usually achieved with geometrical transformations, such as translation, rotation, and rescale applied to the already labeled samples \cite{cite:global-aug,rgb_ral_aug,Yang_2021_CVPR,simclr_chen}.
%

In general, 3D point cloud augmentations \cite{cite:global-aug,cite:LiDAR-Aug} have been much less researched than image augmentation techniques \cite{rgb_ral_aug,simclr_chen,fi14030093,liu2021unbiased}. For example, the aforementioned 3D point cloud augmentations only enrich geometrical features of the training samples, but do not create new scenarios with the previously unseen layout of objects. The lack of modeling realistic class population of the scenes is still a bottleneck of augmentation techniques. This problem can be addressed by augmentation that uses simulated virtual data and scene configurations. However, the effect of such data on training is low due to nonrealistic physical and visual features compared to real data.
\begin{figure}[t]
    \centering
    \includegraphics[width=\linewidth,keepaspectratio]{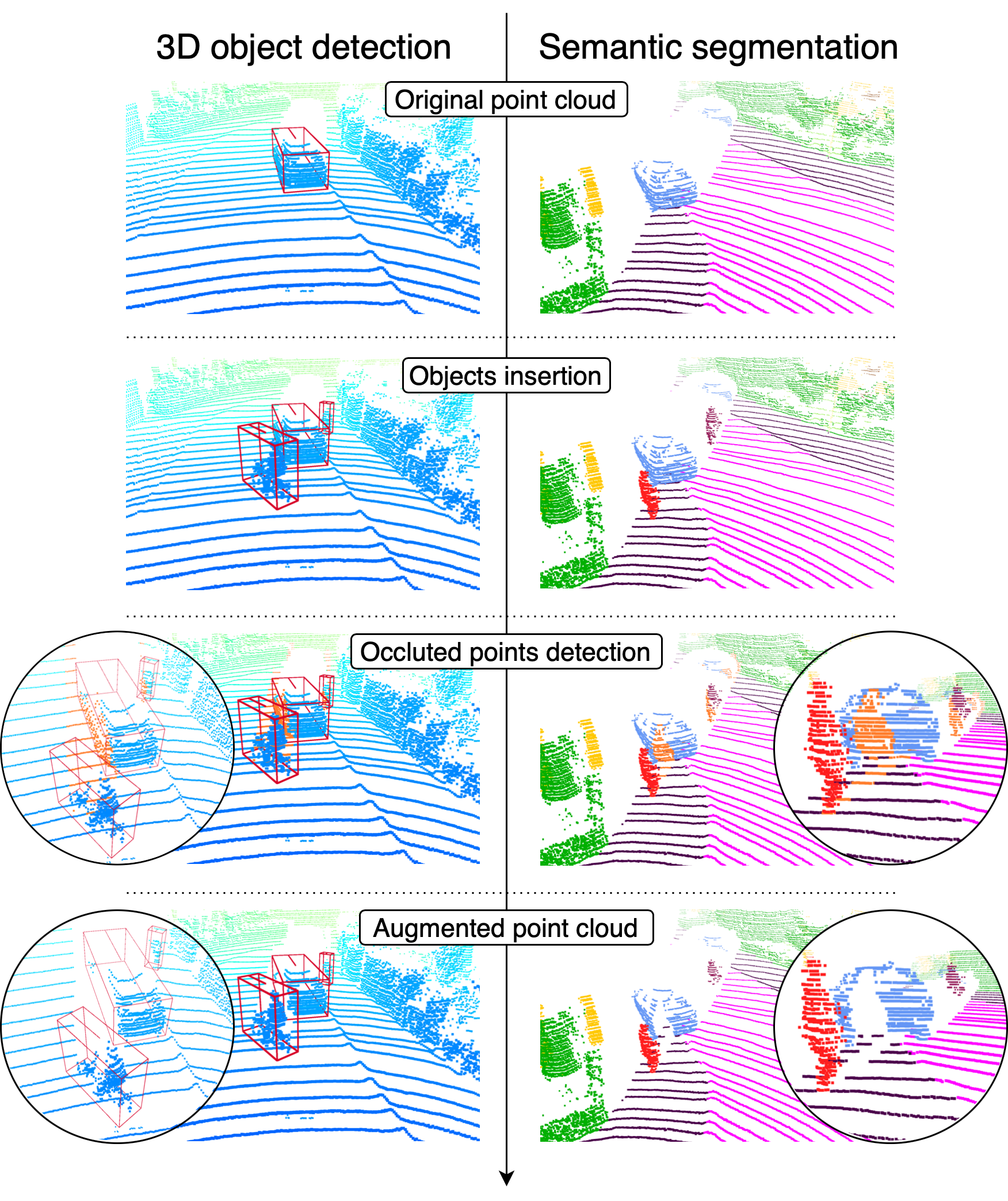}
    \caption{We show examples of our augmentation method in 3D object detection and semantic segmentation. First, we insert objects one by one and then simulate their visibility to model realistic occlusions. Note the details of the scene (circled) and the detection of occluded orange points. After removal, we see the final augmented version of the point cloud in the last row}
    \label{fig:catch_eye}
\end{figure}

We focus on improving the learning of 3D perception networks by enhancing the lidar data in autonomous driving scenarios with data augmentation.
Depth information allows for per-object manipulation when augmenting the point clouds \cite{cite:LiDAR-Aug}. We take advantage of the spatial position of annotated objects and place them in different scenes, while handling occlusions and class-specific inhabitancy.

Our method segments the road and sidewalks for class-specific insertion. Next, the method exploits the bounding boxes of objects to avoid collisions. Compared to state-of-the-art LiDAR-Aug \cite{cite:LiDAR-Aug}, which is suitable only for object detection, our bounding box generation allows augmenting the semantic segmentation datasets and simulates realistic occlusions throughout the spherical projection. The inserted augmentations come from the same dataset and are placed to the same distance, ensuring natural reflection values and point distribution, including ray dropouts. We evaluate the proposed method on tasks of 3D object detection and semantic segmentation. 
Our contribution is twofold:
 \begin{itemize}
     \item We present a new augmentation framework suitable for \emph{both} 3D object detection and semantic segmentation.
     \item We propose a novel way to \emph{model occlusions} and physically consistent insertion of objects for augmentation.
 \end{itemize}
We demonstrate the usefulness of our method on autonomous driving benchmarks and show improvement especially in rarely represented classes. The codes for our method will be publicly available.

%% file: 03-related_work.tex
\section{RELATED WORK}

\subsection{\textbf{Data Augmentation}}
One of the first approaches to augmenting lidar data was GT-Aug, which was published within the 3D detection model SECOND \cite{cite:SECOND_GT-AUG}. GT-Aug adds samples from the ground-truth database, which is precomputed before the training phase. The samples are randomly selected and inserted into the scene as is. If a collision occurs, the added object is simply removed. The visibility and occlusion handling of added scan points or the inserting strategy is not taken into account.
Global data augmentations (Gl-Aug) \cite{cite:global-aug} such as rotation, flip, and scale are commonly used in 3D point-cloud neural networks. These augmentations provide a different geometrical perspective, which supports the neural network with more diversity of training samples. An attempt to automate the augmentation strategy was proposed in \cite{Cheng2020Improving3O}, which narrows the search space based on previous training iterations. The state-of-the-art LiDAR-Aug~\cite{cite:LiDAR-Aug} enriches the training data to improve the performance of the 3D detectors. Additional objects are rendered on the basis of CAD models. Simulations of intensity and raydrops are not discussed in the article. LiDAR-Aug~\cite{cite:LiDAR-Aug} also simulates occlusion between additional objects and the rest of the scene, unlike GT-Aug~\cite{cite:SECOND_GT-AUG}.


\subsection{\textbf{Data Simulators}}
The recent progress in computer vision brought large neural networks with a large number of learnable parameters, often unable to reach saturation point with the size of current training sets. These models require training on a very large number of annotated examples. Commonly used solutions include synthetically generated data~\cite{vacek} or using game simulators such as Grand Theft Auto V, which was used to generate images for the semantic segmentation of ground truth~\cite{cite:GTA}. Some simulators built on Unreal Engine, for example, Carla~\cite{CARLA} are also used in autonomous driving research.
However, the gap between real and synthetic data remains a great challenge~\cite{vacek}. One of the approaches to deal with the difference and portability to the real world is~\cite{Lidar_sim2real,unsupervisedGenerator}, which can produce more realistic lidar data from simulation by learning GAN models.

\subsection{\textbf{3D perception tasks}}

Learning in the lidar point cloud domain poses challenges such as low point density in remote areas, unordered structure of the data, and sparsity due to the sensor resolution. Three common approaches to aggregation and learning the lidar features are voxel-based models \cite{voxelnet,cite:SECOND_GT-AUG}, re-projection of data into 2D structure \cite{xu2020squeezesegv3,Milioto2019RangeNetF}, and point cloud-based models \cite{cite:3Dcylinder,cite:SPVNAS}. To show the ability to generalize, we evaluate our proposed method based on different model feature extractors, as well as on two different tasks of 3D object detection and semantic segmentation task.

One of the key aspects of our approach is placing the object in a realistic position by estimating the road for vehicle and cyclist insertions and the sidewalk for pedestrian insertion. Work \cite{cite:road-segmentation-BEV} designed a fast fully convolutional neural network, which can predict the road from the bird's eye view projection of the scene. However, this approach does not handle occlusions. It does not predict the road behind obstacles, e.g. vehicles. Approaches \cite{cite:ground-seq-1, cite:ground-seq-2} can separate ground from non-ground points, which can even be improved by usage of The Jump-Convolution-Process \cite{cite:JCP}. However, this method filters out all ground points on roads and sidewalks. We need to distinguish them.


%% file: 04-method.tex
\section{method}
Our augmentation method places additional objects into an already captured point cloud. The objects must be placed in adequate locations; therefore, the road and pedestrian area must be estimated (in Subsection \ref{subsec:road-estimation}). The method avoids collisions between additional objects and objects that are in the original point cloud. We analyze overlapping bounding boxes. Therefore, we need to create bounding boxes for semantic datasets that come without object boxes (in Subsection \ref{subsec:object-bbox}). More details on placing additional objects are given in Subsection \ref{subsec:placing-object}. Lastly, the method handles realistic occlusions between objects (in Subsection \ref{subsec:occlusion-handling}). The overview of the proposed method is visualized in figure \ref{fig:overview_v2}.
%
\begin{figure}[h!]
    \centering
    \includegraphics[width=0.99\linewidth, height=0.7\textheight]{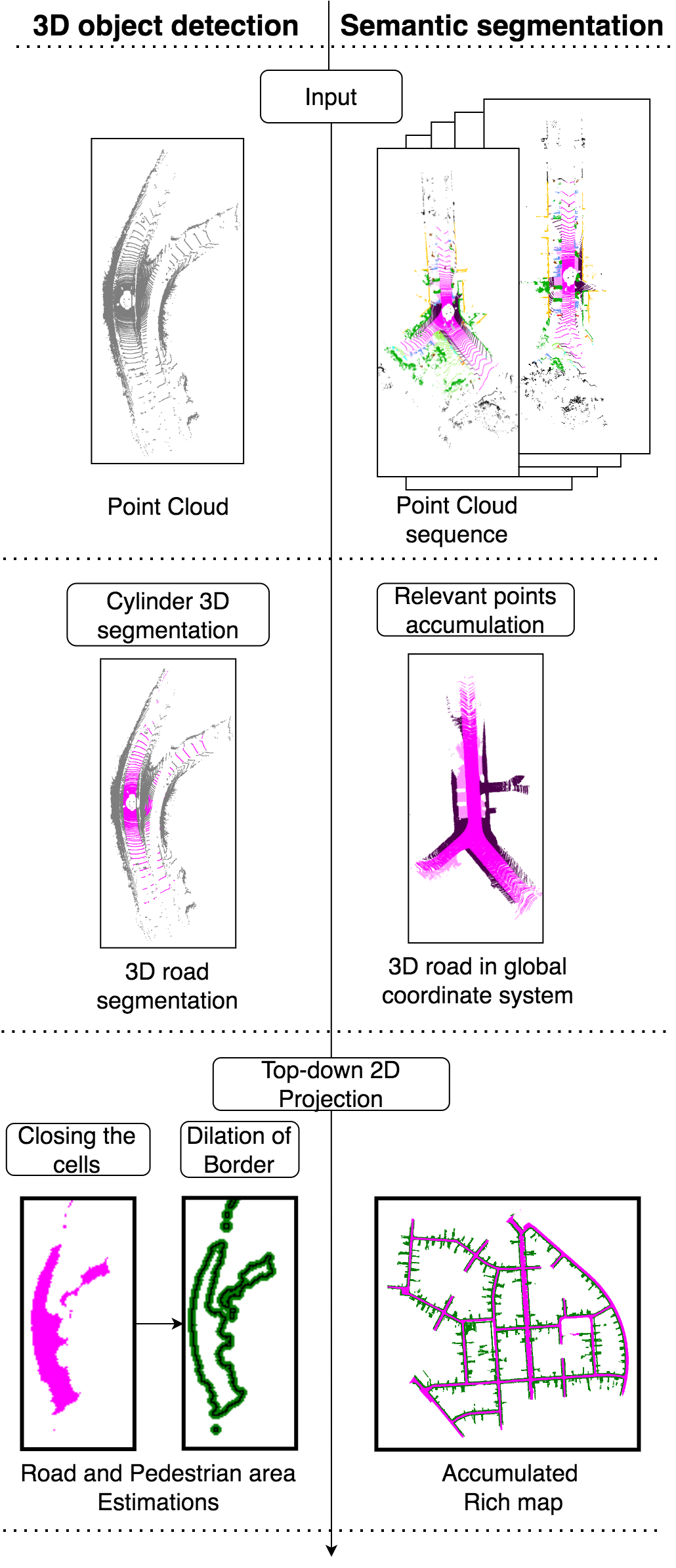}
    \caption{Rich map generating. Road maps are created from points position and labels. Semantic datasets already contains labels of each road point, in case of detection dataset labels are pseudo-labeled by neural network \cite{cite:3Dcylinder}. We then project segmented points into 2D bird's eye view and acquire road and sidewalk map by morphological operations on the 2D projection, namely closing for the road and dilation of road boundary for the sidewalk--pedestrian area.}
    \label{fig:preprocessing}
\end{figure}

\subsection{\textbf{Road estimation}}\label{subsec:road-estimation}
To place the new objects, we need to know where they realistically appear in the scene. This information may be given  by HD maps~\cite{nuscenes,argoverse}, if included in datasets; however, KITTI dataset does not provide them. We estimate valid roads and sidewalk areas for both tasks according to the pipeline described in Fig.~\ref{fig:preprocessing}. First, we pseudo-label 3D points by Cylinder3D~\cite{cite:3Dcylinder}, a state-of-the-art semantic segmentation neural network, which was pre-trained on the SemanticKITTI dataset~\cite{cite:SemanticKitti}. The resulting predictions are then projected onto the 2D lidar $(x,y)$ ground plane, discretized with a cell size resolution of $1\times 1$ meter. Then we divide the space in the scene for the road (cyclist placement) and the sidewalk (pedestrian placement) as follows:

{\bf Road.} To obtain a continuous \textit{road area}, a morphological closing is used on the projection. We use a disk seed with dimension of three.

{\bf Pedestrian area.} The estimate is based on the assumption that pedestrians are supposed to walk along the road border. Cells closer than two pixels from the border of the road estimate are processed and subsequently dilated. We use disk seed with dimension of two.

SemanticKITTI contains poses of each point cloud in sequence. Therefore, road and sidewalk labels can be transformed into a global coordinate system and accumulated in space. The accumulated sequence of road and sidewalk labels leads to a more accurate estimation of the placement areas in the 2D lidar $(x,y)$ ground plane projection. Accumulating multiple scans in one frame densifies the lidar point cloud and naturally reduces the need for morphological operations.



\begin{figure}[t] 
    \centering
    \includegraphics[width=\linewidth]{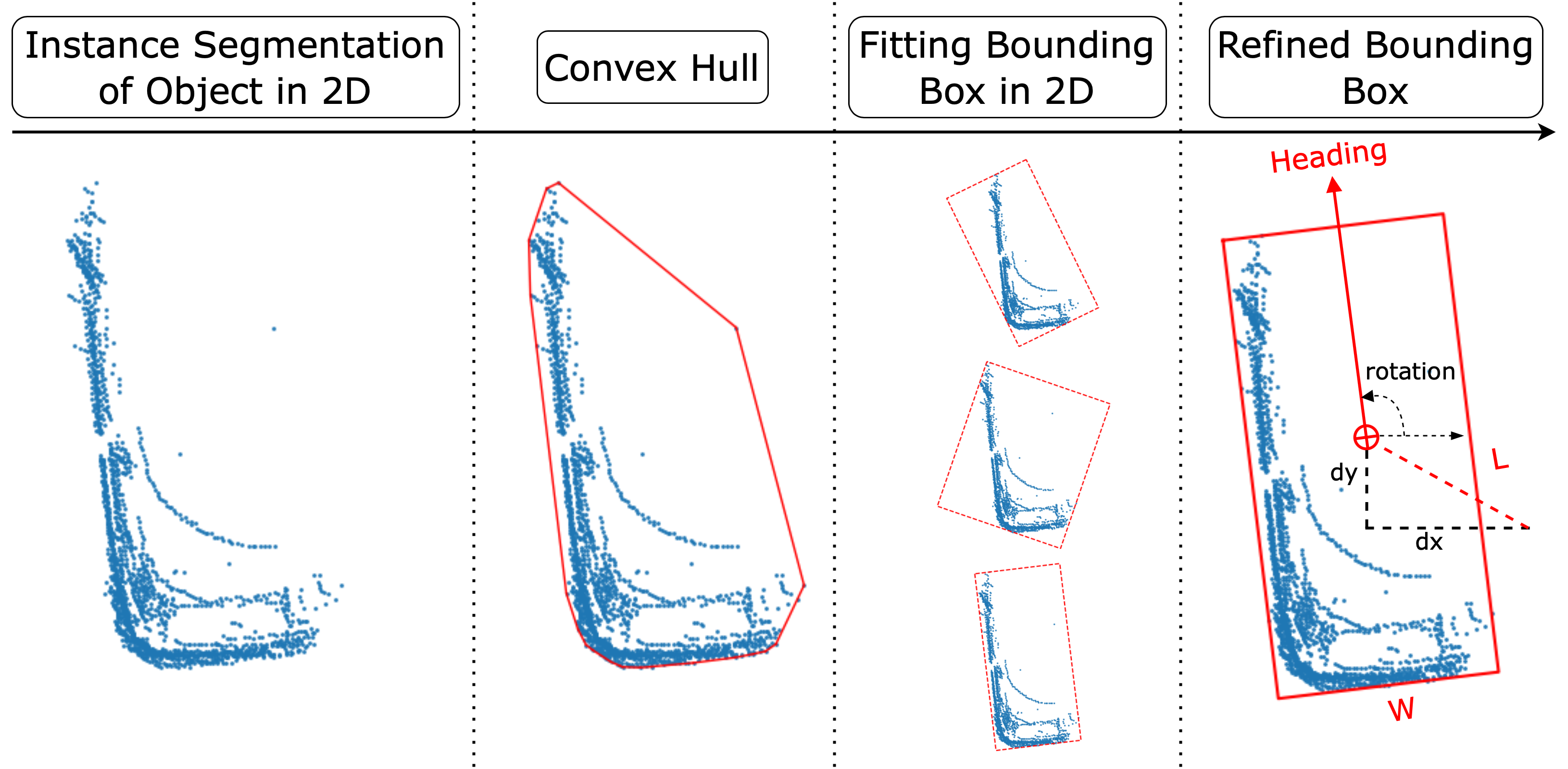}
    \caption{Creation of the bounding box in Bird's Eye View around the car. First, convex hull is constructed around points, then we fit bounding box to estimate position  \textit{x}, \textit{y}, dimensions \textit{length}, \textit{width}, \textit{height}, and orientation \textit{yaw}. The \textit{z} is estimated as if the object touches the road without intersecting it.}
    \label{fig:box_from_points}
\end{figure}

\subsection{\textbf{Creating of bounding boxes}}\label{subsec:object-bbox}

For a collision-free placement of objects, the bounding boxes are required. The bounding box is parameterized by the center coordinates ($x,y,z$), size dimensions ($l,w,h$) and heading angle ($\mathit{yaw}$). For object detection in KITTI dataset, the bounding boxes are already provided as ground-truth labels. However, the SemanticKITTI dataset contains only the semantic label of the class with the instance of the object (each object in one frame has a different instance). We mitigate the absence of the bounding boxes by separating individual objects from the scene based on an instance and estimate bounding boxes, see Fig. \ref{fig:box_from_points}. Modeling the bounding boxes is divided into three steps:

\begin{figure*}[th]
    \centering
    \includegraphics[width=\textwidth]{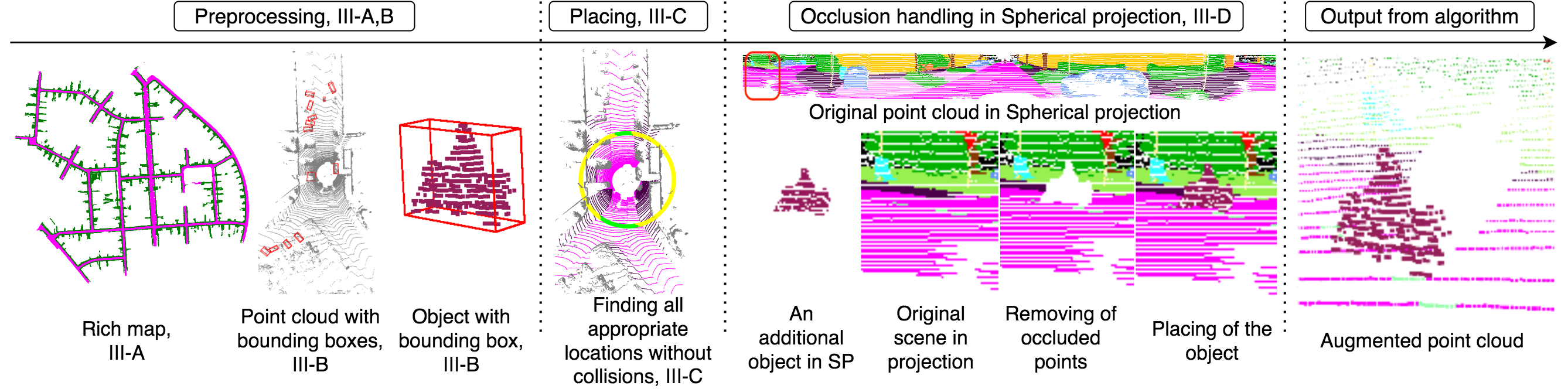}
    \caption{Overview of proposed pipeline. We process the data in order to estimate all possible placements, all bounding boxes in the scene and augmenting objects from different frame. Possible placement of augmenting objects is a conjunction of the same depth as the cut out object (yellow circle) and suitable area from the map of possible insertions (green). 
    Occlusion handling is performed in spherical projection. Result is re-projected to the scene to the 3D augmented point cloud.}
    \label{fig:overview_v2}
\end{figure*}

\textbf{Wrapping.} 
Object-labeled 3D Lidar points are projected to the ground plane. The 2D projected points are wrapped in a convex hull.
    
\textbf{Smallest area.} Assume the convex hull consists of $n$ points. We construct $n-1$ rectangles so that two neighboring points on the convex hull compose one side of the rectangle. The remaining sides of the rectangle are added to achieve the smallest area. 

    
\textbf{Refinement.} Some objects may be represented by too few points. They are scanned at a great distance or are significantly occluded by closer objects. Bounding boxes may also be distorted by occlusions.
We analyze the heights, widths and lengths of the bounding boxes in the KITTI dataset for classes ``Car'', ``Pedestrian'' and ``Cyclists'', which we use in Semantic KITTI. We obtain the distributions for each class and parameter. For each random variable, we calculate the lowest decile. The lowest decile values are the minimal threshold values of the bounding box. The maximal values of bounding boxes are set as the maximal values for the corresponding dimension that occurred in the KITTI data set.
    
    
For bicycle, motorcycle, motorcyclist and truck objects in the SemanticKITTI dataset, we do not have corresponding statistics for bounding box dimensions, since they are not present in KITTI. Therefore, we do not apply any bounding box post-processing in this step for the aforementioned classes.

\subsection{\textbf{Placing of objects}}\label{subsec:placing-object}

For placing of one or multiple objects, we need their bounding box dimensions and yaw angles. Only points within the bounding boxes are used to augment different frames of the dataset. For the semantic segmentation datasets (task), these points are further filtered to have an appropriate label. In the case of the object detection datasets, points that are pseudo-labeled as road or sidewalk are removed to ensure that the cutout point cloud contains only the objects points. 

To maintain the most realistic augmentation, our method places the object at the same distance with the same observation angle, which can be achieved by rotating its point cloud by the vertical z-axis of the frame origin. In this way, realistic object point density and lidar intensity are maintained due to the preserved range between the sensor and the object. It also keeps the same observation angle. Then, we consider the collision-free location of the insertion:

\textbf{Location:} Objects must be fully located on the appropriate surface. For vehicles and cyclists, it is on the road and for pedestrians, it is the pedestrian area. For each appropriate position, the \textit{z} coordinate of the object is adjusted to ensure that the object touches the surface, according to the road prediction level. 

\textbf{Collision avoidance:} At first, the sole bounding box belonging to the object is cut from the scene and placed in the augmented frame on the road level. For the insertion of vehicles and cyclists, the bounding box must not contain any point that is not predicted as road; same for pedestrians and the pedestrian area. Then, we check whether the inserted bounding box overlaps with each of the original boxes from the augmented scene and skip insertion when it does.

\subsection{\textbf{Occlusion handling}}\label{subsec:occlusion-handling}
\begin{algorithm}[H]
	\caption{Occlusion handling}
	\label{alg:occlusion-handling}
	\textbf{Input:} Scene point-cloud, Scene projection, Object point-cloud, Object projection\\
    \textbf{Output:} success, Scene point-cloud 
	\begin{algorithmic}[1]
	    \State sum\_points $\gets$ 0
	    \State success $\gets$ False
		\For {each pixel in object's spherical projection}
			    \If{distance of object is smaller then in scene}
    			        \State Remove scene points in pixel (they are occluded)
    			        \State Add points projected to object s. p. pixel to scene
    			        \State sum\_points $\gets$ sum\_points $+$ nbr of added points
				\EndIf
		\EndFor
	    \If{sum\_points $>$ minimal point for class}
	        \State success $\gets$ True
        \EndIf    
        \State\Return{success, Scene}
	\end{algorithmic} 
\end{algorithm}
\textbf{Data projection:} The occlusion handling uses a spherical projection, similarly to~\cite{xu2020squeezesegv3}, to solve realistic visibility after the additional object is placed. The spherical projection stores information about the minimal distance between the sensor and the points that were projected to the corresponding pixel. To correct the holes in the object, the projection is morphologically closed by a rectangular seed of dimension $5\times3$ (5 rows and 3 columns). The pixels closed by the seed are assigned the depth computed from the neighboring pixels as an average of the depths in that seed area.
Morphological closing is computed separately for the scene and object.

\textbf{Removing occluded points:} The algorithm goes through every pixel in the spherical projection. Every pixel contains information about the distance of the point. All scene points more distant than the inserted point are removed, since they would be naturally occluded by the added object.
as they are occluded by the placed object. Consequently, all object points, which were projected in the same pixel, are added to the scene point cloud. The procedure ends if a sufficient number of points is added to the scene. A pseudocode of the algorithm is shown in Algorithm \ref{alg:occlusion-handling}.
%

%% file: 05-experiments.tex
\section{Experiments}

\subsection{\textbf{Datasets and perception tasks}}

\textbf{3D object detection:} We use the KITTI 3D object detection benchmark. The data set consists of 7,481 training scenes and 7,518 testing scenes with three object classes: ``car'', ``pedestrian'', and ``cyclist''.

The test labels are not accessible, and access to the test server is limited. Therefore, we followed the methodology proposed by~\cite{cite:LiDAR-Aug} and divided the training data set into training and validation parts, where the training set contains 3,712 and the validation 3,769 lidar samples~\cite{cite:trainKitti}. The split of the dataset into training and validation was made consistent with standard KITTI format, i.e. with regard to avoid having similar frames and scenes in both sets. The evaluation was carried out on a validation set, where the labels are available, as was done in \cite{cite:LiDAR-Aug,cite:KITTI}.

For object detection, we consider only the two more rare and challenging classes, i.e. pedestrian and cyclist. From our point of view, the datasets provide a sufficient number of examples of cars for training.

A metric for conducting an evaluation is the standard average precision (AP) of 11 uniformly sampled recall values. We use the IoU threshold $50\%$, true positive predictions are considered bounding boxes with ground-truth overlaps greater than $50\%$. We denote AP for ``Pedestrian'' as \textbf{AP}$_{\text{Ped}}$\textbf{50}($\%$) and \textbf{AP}$_{\text{Cyc}}$\textbf{50}($\%$) for ``Cyclist''. The difficulties of the predictions are divided on the basis of the sizes of the bounding box, occlusion, and truncation into ``Easy'', ``Moderate'' and ``Hard'' as required by the \cite{cite:KITTI} benchmark.

\textbf{Semantic segmentation:} We use the SemanticKITTI \cite{cite:SemanticKitti} benchmark. The dataset is an extension of the original KITTI~\cite{cite:KITTI} benchmark with dense point-wise annotations provided for each $360^\circ$ field-of-view frame. In general, the dataset offers 23,201 3D scans for training and 20,351 for testing. The training data set was divided into training and validation parts with 19 annotated classes.

The intersection-over-union, \mbox{$\text{IoU} = {\text{TP}}/({\text{TP}+\text{FP}+\text{FN}})$}, was used for comparison. Performance is evaluated for each individual class, as well as the average (mIoU) for all classes.

\subsection{\textbf{3D perception models}}

We tested the augmented data on two \emph{3D object detection} models, each based on a different type of feature extractor backbone. \textit{PV-RCNN} \cite{cite:PV-RCNN} is a 3D object detection model that combines a 3D voxel convolutional neural network with a pointnet-based set abstraction approach~\cite{pointnet++}. The second is PointPillar~\cite{cite:PointPillars}, which encodes the point cloud in vertical pillars. The pillars are later transformed into 3D pseudo-image features. 

For \emph{segmentation task}, we use the SPVNAS~\cite{cite:SPVNAS} multiclass detector. The neural network achieves significant computation reduction due to the sparse Point-Voxel convolution and holds the fourth place on the competitive SemanticKITTI leaderboard and the second among those with available implementation.

Each neural network was set to the default parameters proposed by the authors of the architectures, with its performance reported on KITTI 3D benchmark and SemanticKITTI. We trained each neural network three times for object detection and five times for semantic segmentation. Average performance was considered as the final score of the method.


\subsection{\textbf{Augmentations}}

All augmentations were trained with the same hyperparameters to ensure a fair comparison between methods. The approach of GT-Aug was performed with information of the precomputed planes, which is an approximation of the ground from the KITTI dataset. This step should ensure that the inserted objects lie on the ground. For our proposed augmentation method, we choose to add objects with a zero-occlusion KITTI label only (Easy). Some cases are naturally transformed into other difficulties (Moderate and Hard) by newly created occlusions.
%


For global augmentation of the scenes, we used uniformly distributed scaling of the scene in the range $[0.95, 1.05]$, rotation around the z-axis (vertical axis) in the range $[-45^{\circ}, 45^\circ]$ and random flipping over the x-axis from the point cloud as in \cite{cite:global-aug,cite:LiDAR-Aug}.

The maximum number of added objects in semantic segmentation was set to 10 per scene, and object class is selected randomly each time of the insertion.
In Table \ref{tab:sem} we can see the number of objects added per scene and the minimal number of points for insertion. 
\begin{table}[h]
    \centering
    \caption{Number of insertions per class on Semantic KITTI dataset}
    \label{tab:sem}
    \begin{tabular}{l|c|c}
    \toprule
    Class &  Min number of points & Added objects per scene\\
    \midrule
    Bicycle & 10 & 1.67\\
    Motorcycle & 10 & 1.66\\
    Truck & 40 & 1.59\\
    Person & 20 & 1.56\\
    Bicyclist & 30 & 1.66\\
    Motorcyclist & 30 & 1.66\\
    \bottomrule
\end{tabular}
\end{table}

\begin{table}[b]
\centering
\caption{Object detection results with PV-RCNN. Our method achieves the best results in categories ``pedestrian'' and ``easy cyclists''. (mc) abbreviates multiclass}
\begin{tabular}{lccccccc} 
\toprule
\multirow{1}{*}{} 
& \multicolumn{3}{c}{$\textbf{AP}_{\text{Ped}}$ \textbf{50}($\%$)}  
& \multicolumn{3}{c}{$\textbf{AP}_{\text{Cyc}}$ \textbf{50}($\%$)} \\ 
Method & Easy & Mod & Hard & Easy & Mod & Hard\\
\midrule
w/o Object-Aug &65.92 &59.14 &54.51 &76.80 &59.36 &56.61\\
GT-Aug \cite{cite:SECOND_GT-AUG} &65.69 &59.33 &54.78 &88.30 &\textbf{72.55} &\textbf{67.79}\\
LiDAR-Aug \cite{cite:LiDAR-Aug} &65.05 &58.90 &55.52 &N/A&N/A&N/A\\
Real3D-Aug (mc) &\textbf{73.57} &\textbf{66.55} &\textbf{62.17} &\textbf{92.69} &65.06 &63.43\\
\hline
\end{tabular}
\label{tab:results3D_obj}
\end{table}

\subsection{\textbf{Evaluation}}

%
\begin{table*}[t!]
\caption{Semantic segmentation on SemanticKITTI. Comparison of our method with the global augmentation baseline. Both methods use SPVNAS~\cite{cite:SPVNAS} neural network. The reported results are averaged over five runs. The augmented categories are typeset in \textit{italics}. We observe a performance gain in each of them except for one: trucks. Improvement is especially notable in the motorcyclist class, which contains only a few training examples in the dataset with only global augmentations.}
    \label{tab:seg}
    \centering
        \setlength{\tabcolsep}{2.5pt}
        \begin{tabular}{lrrrrrrrrrrrrrrrrrrrr}
        
        \toprule
        {} & {{\textbf{mIoU}}} & {\rotatebox[origin=l]{90}{car}} & {\rotatebox[origin=l]{90}{\textit{bicycle}}} & {\rotatebox[origin=l]{90}{\textit{motorcycle}}} & {\rotatebox[origin=l]{90}{\textit{truck}}} & {\rotatebox[origin=l]{90}{other-vehicle}} & {\rotatebox[origin=l]{90}{\textit{person}}} & {\rotatebox[origin=l]{90}{\textit{bicyclist}}} & {\rotatebox[origin=l]{90}{\textit{motorcyclist}}} & {\rotatebox[origin=l]{90}{road}} & {\rotatebox[origin=l]{90}{parking}} & {\rotatebox[origin=l]{90}{sidewalk}} & {\rotatebox[origin=l]{90}{other-ground}} & {\rotatebox[origin=l]{90}{building}} & {\rotatebox[origin=l]{90}{fence}} & {\rotatebox[origin=l]{90}{vegetation}} & {\rotatebox[origin=l]{90}{trunk}} & {\rotatebox[origin=l]{90}{terrain}} & {\rotatebox[origin=l]{90}{pole}} & {\rotatebox[origin=l]{90}{traffic-sign}} \\

        \midrule
         w/o Obj-Aug  &     60.62 &  95.47 &    29.64 &       58.16 &  \textbf{64.22} &          47.69 &   66.24 &      79.14 &          0.04 &  \textbf{93.06} &    \textbf{48.52} &     \textbf{80.20} &          1.72 &     \textbf{89.75} &  \textbf{58.67} &       87.88 &  67.07 &    73.40 &  63.51 &         47.34 \\
        Real3D-Aug &     \textbf{62.76} &  \textbf{95.93} &    \textbf{44.13} &       \textbf{73.41} &  49.24 &          \textbf{48.43} &   \textbf{70.34} &      \textbf{85.45} &          \textbf{12.01} &  92.84 &    45.66 &     79.66 &          \textbf{2.91} &     89.36 &  56.96 &       \textbf{89.18} &  \textbf{67.61} &    \textbf{76.72} &  \textbf{63.73} &         \textbf{48.88} \\
        
        \bottomrule
        \end{tabular}
\end{table*}

We compare our method (Real3D-Aug) with copy-and-paste augmentation (GT-Aug) \cite{cite:SECOND_GT-AUG} and with state-of-the-art LiDAR-Aug augmentation \cite{cite:LiDAR-Aug}.
In the Real3D-Aug multiclass (mc), we added $4.7$ pedestrians and $6.7$ cyclists on average per scene. All methods were trained with global augmentations \cite{cite:global-aug} if not stated otherwise.



In Table \ref{tab:results3D_obj} we show the results of LiDAR-Aug with PV-RCNN. The numbers are taken from the original paper due to the unpublished codes and the lack of technical details about their CAD model and ray-drop characteristic. In the original article, LiDAR-Aug was trained under unknown hyperparameters and was not applied to the cyclist category. Our method surpasses the LiDAR-Aug in pedestrian class by a large margin despite all the difficulties. Both GT-Aug and Real3D-Aug achieve a significant improvement in performance. Real3D-Aug achieves a significant improvement with PV-RCNN in the pedestrian class, where we achieve $15.4\%$, $10.96\%$, and $7.87\%$ improvement in Easy, Moderate, and Hard difficulty, and GT-Aug achieves $7.52\%$, $3.74\%$, and $0.48\%$ improvement compared to the model without (w/o) any object augmentations.

In the semantic segmentation task, we achieve an increase of $2.14$ in mean IoU compared to the common augmentation technique \cite{cite:global-aug}, see Table \ref{tab:seg}. We observe an increase in IoU of all classes added, except for the truck category. Our method also increases the car category (not augmented) since we add more negative examples to other similar classes. We are not comparing with GT-Aug~\cite{cite:SECOND_GT-AUG} and LiDAR-Aug~\cite{cite:LiDAR-Aug} in the semantic segmentation task. The aforementioned methods were not designed for segmentation, whereas our method allows for augmenting both tasks.

\subsection{\textbf{Ablation study of object detection}}
In Tables \ref{tab:abl-results3D-PointPillar} and \ref{tab:abl-results3D-PVRCNN} we show the influence of adding a single object to the scene in comparison to GT-Aug. Each configuration is named after the added class, and the lower index indicates the average number of objects added per scene. We can see that, in the case of PointPillar, adding only one class decreases performance on the other classes. We suspect that it is conditioned by similarities between classes. For example, pedestrians and bicycles are simultaneously present in the class ``cyclist''. Therefore, it is beneficial to add both classes simultaneously. In case of PV-RCNN, the addition of one class improves the performance of both.

\begin{table}
\caption{Real3D-Aug Object detection results with PointPillar architecture based on number of inserted classes.}
\centering
\begin{adjustbox}{width=\columnwidth,center}
\begin{tabular}{lcccccc} 
\toprule
\multirow{2}{*}{Augmentation} 
& \multicolumn{3}{c}{$\textbf{AP}_{\text{Ped}}$ \textbf{50}($\%$)}  
& \multicolumn{3}{c}{$\textbf{AP}_{\text{Cyc}}$ \textbf{50}($\%$)} \\ 
& Easy & Mod & Hard & Easy & Mod & Hard\\
\midrule
GT-Aug    &54.52 &49.04 &45.49 &\textbf{77.64} &\textbf{61.30} &\textbf{58.15}\\
Real3D-Aug (Ped$_1$)    &\textbf{55.72} & 51.30 &47.47 &46.33 &33.84 &32.47\\
Real3D-Aug (Cyc$_1$)    &46.87 &44.17 &41.77 &72.65 &52.71 &49.04\\
Real3D-Aug (mc) & 55.50 & \textbf{52.00} & \textbf{49.03} & 76.82 & 52.74 & 50.18\\
\bottomrule
\end{tabular}
\end{adjustbox}
\label{tab:abl-results3D-PointPillar}
\end{table}

\begin{table}
\caption{Real3D-Aug object detection results with PV-RCNN based on number of inserted classes.}
\centering
\begin{adjustbox}{width=\columnwidth,center}
\begin{tabular}{lcccccc} 
\toprule
\multirow{2}{*}{Augmentation} 
& \multicolumn{3}{c}{$\textbf{AP}_{\text{Ped}}$ \textbf{50}($\%$)}  
& \multicolumn{3}{c}{$\textbf{AP}_{\text{Cyc}}$ \textbf{50}($\%$)} \\ 
& Easy & Mod & Hard & Easy & Mod & Hard\\
\midrule
GT-Aug &65.69 &59.33 &54.78 & 88.30 &\textbf{72.55} &\textbf{67.79}\\
Real3D-Aug (Ped$_1$) & 70.96 & 66.63 & 61.14 &78.97 &63.47 &57.31\\
Real3D-Aug (Cyc$_1$)  &65.63 &59.14 &57.47 &82.79 &63.69 &62.39\\
Real3D-Aug (mc)  & \textbf{73.57} & \textbf{66.55} & \textbf{62.17} & \textbf{92.69} & 65.06 & 63.43\\
\bottomrule
\end{tabular}
\end{adjustbox}
\label{tab:abl-results3D-PVRCNN}
\end{table}




%% file: 07-conclusion.tex
\section{CONCLUSION}

We propose an object-centered point cloud augmentation technique for 3D detection and semantic segmentation tasks. Our method improves performance on important and rarely occurring classes, e.g. pedestrian, cyclist, motorcyclist, and others. The method outperforms state-of-the-art on well-established autonomous driving benchmarks. Our method is self-contained and requires only 3D data. All augmentations can be preprocessed, so it does not increase time during training. One way to further improve the method is to incorporate more informative selection of placements based on the uncertainty of the detection model.